\documentclass[twoside,11pt]{article}

\usepackage[preprint]{jmlr2e}
\usepackage[utf8]{inputenc} 
\usepackage[T1]{fontenc}    
\usepackage{hyperref}       
\usepackage{url}   
\usepackage[ruled,longend,linesnumbered]{algorithm2e}
\usepackage{float}
\usepackage{multirow}
\usepackage{booktabs}       
\usepackage{amsfonts}    \usepackage{rotating}
\usepackage{nicefrac}       
\usepackage{microtype}  
\usepackage{graphicx}
\usepackage{tabularx}
\usepackage{subfigure}
\usepackage[dvipsnames]{xcolor}
\usepackage{graphicx}
\usepackage{amsmath}
\usepackage{xcolor}
\usepackage{colortbl}
\usepackage{enumitem}
\usepackage{caption}
\usepackage{listings}
\usepackage{appendix}
\usepackage{chngcntr}
\usepackage{minitoc}

\newcolumntype{C}{>{\centering\arraybackslash}X}

\hypersetup{
    colorlinks = true,
    allcolors = {Tan},
    linkbordercolor = {white},
}

\newenvironment{tightcenter}{%
  \setlength\topsep{0pt}
  \setlength\parskip{0pt}
  \begin{center}
}{%
  \end{center}
}

\ShortHeadings{Interactive 3D Medical Image Segmentation with SAM 2}{Chuyun Shen et al.}
\firstpageno{1}

\begin{document}

\doparttoc 
\faketableofcontents 

\title{Interactive 3D Medical Image Segmentation with SAM 2}

\author{\name Chuyun Shen\footnotemark[1] \footnotemark[2]  \email cyshen@stu.ecnu.edu.cn \\
    \addr School of Computer Science and Technology \\
    East China Normal University\\
    Shanghai 200062, China
    \AND
    \name Yifei Huang\footnotemark[2] \email hyifei\_yof@163.com\\
    \addr School of Software Engineering\\
    East China Normal University\\
    Shanghai 200062, China
    \AND
    \name Wenhao Li \email whli@tongji.edu.cn\\
    \addr School of Software Engineering\\
    Shanghai Research Institute for Intelligent Autonomous Systems\\
    Tongji University\\
    Shanghai 200092, China
    \AND
    \name Yuhang Shi \email yuhang.shi@cri-united-imaging.com \\
    \addr Shanghai United Imaging Healthcare Advanced Technology \\ Research Institute Co., Ltd. \\
    Shanghai 201807, China
    \AND
    \name Xiangfeng Wang \email xfwang@cs.ecnu.edu.cn \\
    \addr School of Computer Science and Technology\\
    East China Normal University\\
    Shanghai 200062, China
}

\renewcommand{\thefootnote}{\fnsymbol{footnote}}
\footnotetext[1]{This work was finished during an internship at Shanghai United Imaging Healthcare Advanced Technology.}
\footnotetext[2]{These authors contributed equally to this work.}

\maketitle

\begin{abstract}
Interactive medical image segmentation (IMIS) has shown significant potential in enhancing segmentation accuracy by integrating iterative feedback from medical professionals. 
However, the limited availability of enough 3D medical data restricts the generalization and robustness of most IMIS methods.
The Segment Anything Model (SAM), though effective for 2D images, requires expensive semi-auto slice-by-slice annotations for 3D medical images. 
In this paper, we explore the zero-shot capabilities of SAM 2, the next-generation Meta SAM model trained on videos, for 3D medical image segmentation. 
By treating sequential 2D slices of 3D images as video frames, SAM 2 can fully automatically propagate annotations from a single frame to the entire 3D volume.
We propose a practical pipeline for using SAM 2 in 3D medical image segmentation and present key findings highlighting its efficiency and potential for further optimization.
Concretely, numerical experiments on the BraTS2020 and the medical segmentation decathlon datasets demonstrate that SAM 2 still has a gap
with supervised methods but can narrow the gap in specific settings and organ types, significantly reducing the annotation burden on medical professionals. 
Our code will be open-sourced and available at \href{https://github.com/Chuyun-Shen/SAM\_2\_Medical\_3D}{https://github.com/Chuyun-Shen/SAM\_2\_Medical\_3D}.

\end{abstract}

\section{Introduction}\label{sec:intro}

\begin{figure}[htb!]
    \centering
    \includegraphics[width=0.9\linewidth]{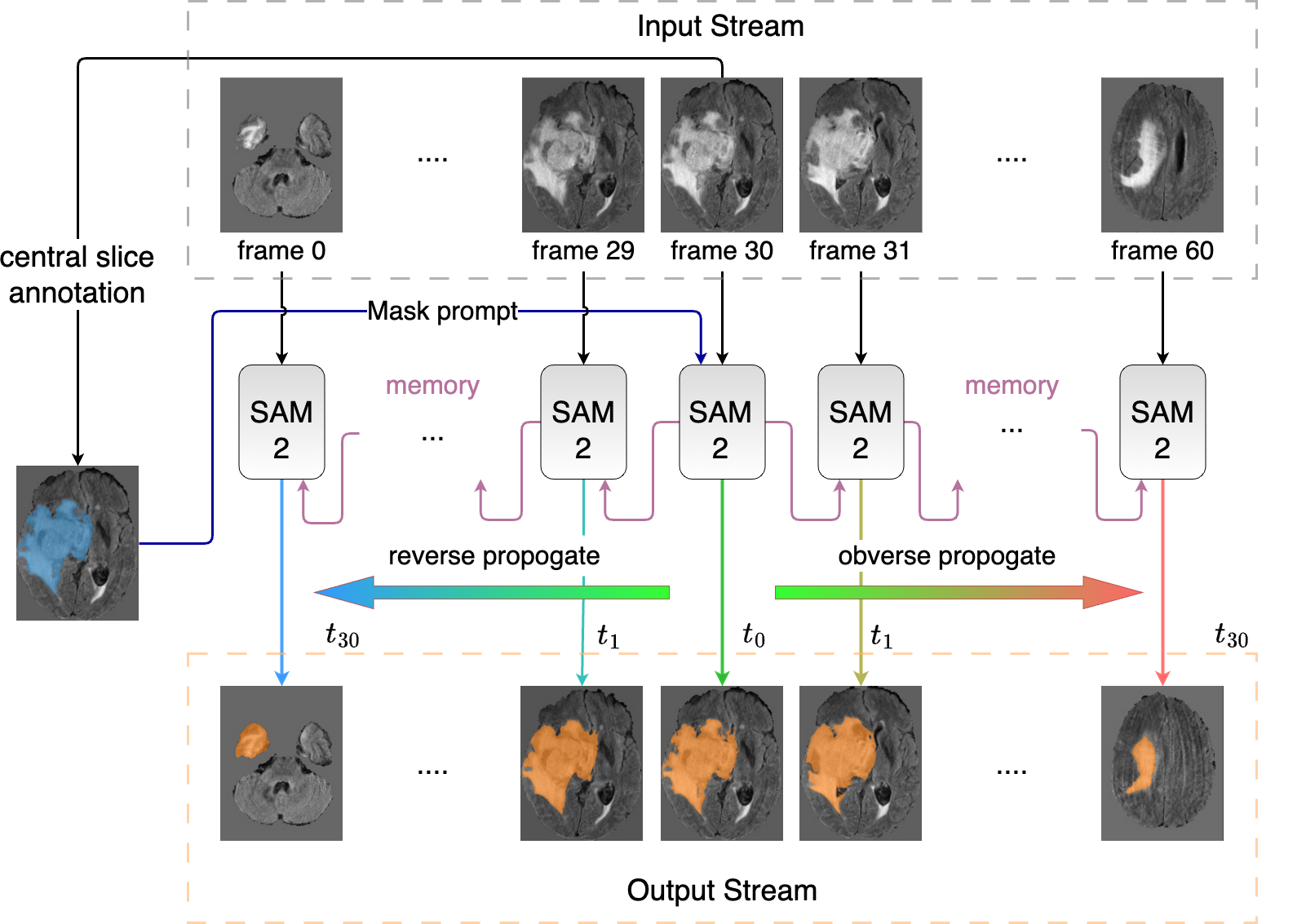}
    \caption{Pipeline Diagram: Utilizing Sam 2 for Propagating Slice Annotations for 3D Interactive Medical Image Segmentation. The central slice first needs to be segmented by a 2D segmentation algorithm or annotated by a human expert either through manual labeling or using an interactive semi-automatic algorithm. SAM 2 inputs the mask prompt and then predicts all other slices sequentially in both directions, ultimately obtaining annotations for all slices.}
    \label{fig:sam2_3d}
\end{figure}

\begin{figure}[htb!]
    \centering
    \includegraphics[width=\linewidth]{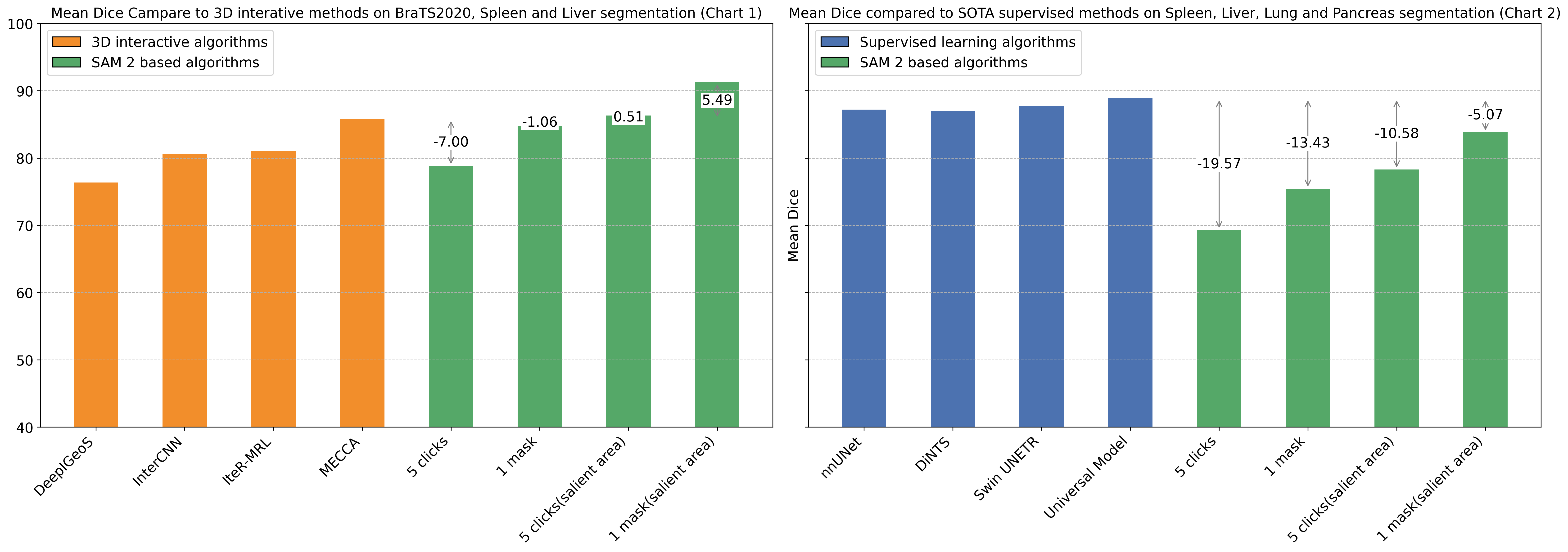}
    \caption{Caparision with 3D interactive methods and supervised methods. the orange bars represent 3D interactive algorithms, which typically handle 3D images by resizing. The blue bars denote supervised learning algorithms, which usually process 3D images using patches. The green bars signify algorithms based on SAM 2 segmentation. In this context, "5 clicks" refers to interactively clicking on five points on the central 2D image using SAM, one point per round, to generate 2D slice annotations, which are then propagated to the 3D image. "1 mask" indicates providing SAM 2 with the ground truth mask of the central 2D image, which is then propagated to the 3D image. "Salient area" refers to results tested only on slices with more than 256 foreground points. The bidirectional arrows indicate the difference in dice score between SAM 2-based algorithms and the optimal algorithms. Chart 1 compares the dice scores of 3D interactive algorithms and SAM 2 on the BraTS2020, Spleen, and Liver datasets, while Chart 2 compares the dice scores of supervised algorithms and SAM 2 on the Spleen, Liver, Lung, and Pancreas datasets.}
    \label{fig:compare_mean_dice}
\end{figure}

\begin{figure}[htbp!]
    \centering
    \includegraphics[width=0.9\linewidth]{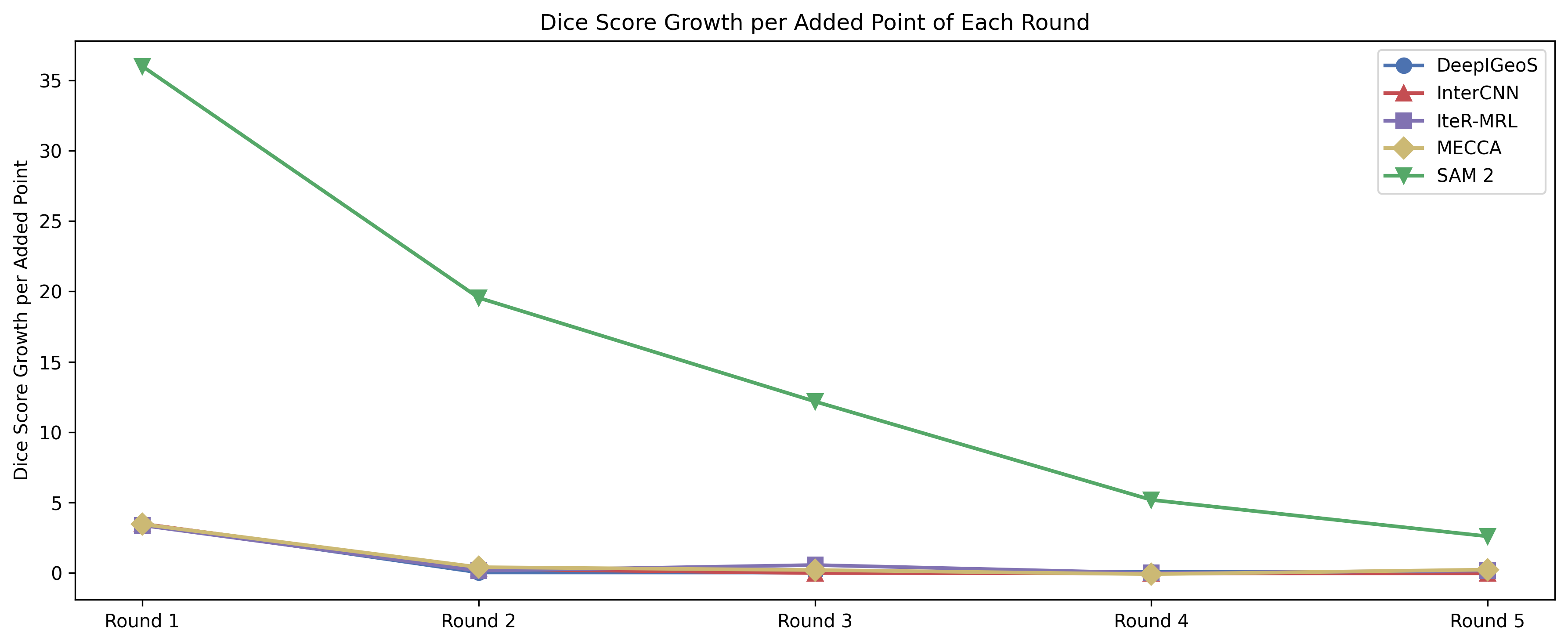}
    \caption{Dice Score Growth per Added Point of Each Round: On the BraTS2020 benchmark, we evaluated how much the average dice score improves per additional point in each round for different interactive algorithms. The interactive methods used by the four algorithms—DeepIGeoS, InterCNN, IteR-MRL, and MECCA—select 25 points in the first round on the 3D medical image, followed by 5 additional points per round. In contrast, our pipeline with SAM 2 adds one point per round.}
    \label{fig:growth_per_added_point}
\end{figure}

Medical image segmentation (MIS)~\citep{unet,nnunet,nnformer,swin-unet} poses distinct challenges compared to natural images due to the diverse modalities, intricate anatomical structures, unclear and complex object boundaries, and varying object scales involved~\citep{sharma2010automated,hesamian2019deep,huang2023segment}. 
Thus, the interactive medical image segmentation (IMIS) paradigm has garnered significant attention for substantially improving performance over conventional methods~\citep{xu2016deep,deepcut,scribblesup,polygon-rnn,deepigeos,seednet,itermrl,boundary-aware,mecca}.

IMIS reimagines MIS as a multi-stage, human-in-the-loop process, where medical professionals provide iterative feedback—such as marking critical points, delineating boundaries, or defining bounding boxes—to refine model outputs. 
This iterative feedback loop allows the model to integrate expert knowledge and progressively enhance segmentation accuracy.
However, the limited availability of medical data restricts most IMIS methods to a few datasets and segmentation tasks, resulting in poor generalization and robustness.

The \textit{Segment Anything Model} (SAM)\citep{sam} has shown exceptional effectiveness in interactive segmentation for natural images and, more recently, medical images, thanks to its prompt-based, zero-shot generalization capabilities
~\citep{ji2023sam,ji2023segment,mohapatra2023brain,deng2023segment,zhou2023can,he2023accuracy,mazurowski2023segment,ma2023segment,cheng2023sam,zhang2023segment,roy2023sam,huang2023segment,mattjie2023exploring}. 
Despite this, SAM's training on 2D natural images presents a significant mismatch with the 3D nature of medical imaging modalities like CT, MRI, and PET. 
Current SAM-based tools require laborious slice-by-slice annotations, even for similar slices, which is impractical in clinical settings.

Fortunately, SAM 2~\citep{ravi2024sam2}, the next generation of Meta SAM trained on videos, offers a promising solution. 
SAM 2 can segment entire videos based on annotations from a \textit{single} frame, utilizing interactions (clicks, boxes, or masks) on any frame to predict spatiotemporal masks, or `masklets.' 
Different slices of 3D medical images are sequentially scanned and stacked over time, allowing 3D medical images to be naturally regarded as videos.
This naturally raises the following question:

\begin{tightcenter}
   \textit{Can SAM 2 segments 3D medical image based solely on 2D interactive feedbacks in a zero-shot manner?} 
\end{tightcenter}



If we can get an affirmative answer, this paradigm shift could enable researchers to focus on automatic segmentation for single 2D images, thus significantly reducing the amount of expert annotation required for 3D interactive segmentation. 
This paper attempts to preliminarily answer this question from an experimental perspective and has obtained some dialectical observations.
Concretely, we propose a simple and practical pipeline (Shown in Fig.\ref{fig:sam2_3d}) to enable the use of SAM 2 for 3D medical images, evaluate SAM 2's zero-shot performance on the Brats and some MSD datasets, 
and get the following key observations:

\noindent (1) The experimental results suggest that SAM 2, in a zero-shot manner, still has a gap with supervised methods but can narrow the gap in specific settings and organ types (shown in Fig.\ref{fig:compare_mean_dice}). Further optimization and refinement of the medical 3D images is necessary.

\noindent (2) SAM 2's efficiency in utilizing interactive feedback significantly surpasses that of other 3D interactive medical image segmentation algorithms. (shown in Fig.\ref{fig:growth_per_added_point})

\noindent\textbf {Remark.}
Since the release of SAM 2, two works have explored its application in medical image segmentation.  
\citet{dong2024segment} introduce SAM 2 for 3D Medical Imaging by treating each slice as a frame and leveraging a memory bank for prediction propagation.
They conduct an extensive evaluation of SAM 2 using 18 diverse medical imaging datasets, demonstrating its performance in both single-frame 2D segmentation and multi-frame 3D segmentation.
They also identify key strategies for enhancing SAM 2's segmentation accuracy, including selecting the center slice of the object of interest, utilizing bidirectional propagation, and preferring the first predicted mask over the most confident one.
Another notable work, the MedSAM-2 Framework \citep{zhu2024medical}, represents the first SAM-2-based model for medical image segmentation, addressing both 2D and 3D tasks. MedSAM-2 incorporates the Confidence Memory Bank and Weighted Pick-up strategy, surpassing state-of-the-art models across 15 benchmarks and 26 tasks, thereby demonstrating superior generalization and performance.
In contrast to these studies, our work does not explore different modes; rather, it adopts settings specifically tailored for medical imaging, akin to the optimal strategy mentioned in \citet{dong2024segment}. We also discussed and investigated the feasibility of interactive annotation based on SAM 2 on 2D slices, subsequently propagating these annotations to 3D images. Additionally, we compared the accuracy of this method with traditional 3D interactive medical image segmentation algorithms and supervised learning algorithms, highlighting the gap in performance with them.

\section{Related Work and Preliminaries}

\subsection{3D Interactive Medical Image Segmentation}

In recent years, deep learning-based interactive medical image segmentation (IMIS) methods have garnered significant interest.
\citet{xu2016deep} proposed using convolutional neural networks (CNNs) for interactive image segmentation. 
Techniques like DeepCut~\citep{deepcut} and ScribbleSup~\citep{scribblesup} leverage weak supervision to develop interactive segmentation approaches. Additionally, DeepIGeoS~\citep{deepigeos} incorporates a geodesic distance metric to create a hint map for improved segmentation accuracy.

The sequential nature of the interactive segmentation process makes it well-suited for reinforcement learning (RL). Polygon-RNN~\citep{polygon-rnn} addresses this by treating segmentation targets as polygons and iteratively selecting polygon vertices via a recurrent neural network (RNN). Similarly, Polygon-RNN+~\citep{polygon-rnn++} employs RL to enhance vertex selection further. SeedNet\citep{seednet} takes a distinct approach by developing an RL model for expert interaction generation, enabling the acquisition of simulated interaction data at each stage of the segmentation process.
IteR-MRL~\citep{itermrl} and BS-IRIS~\citep{boundary-aware} frame the dynamic interaction process as a Markov Decision Process (MDP), utilizing multi-agent RL models for image segmentation. Building on IteR-MRL, MECCA~\citep{mecca} introduces a confidence network to address the common issue of "interactive misunderstanding" in RL-based IMIS techniques and to enhance the utilization of human feedback.  Additionally, \citet{marinov2023deep} provides a thorough review of the IMIS domain.
These advancements underscore the potential of deep learning and reinforcement learning in revolutionizing interactive medical image segmentation, leading to more accurate and efficient segmentation techniques.

\subsection{Segment Anything Model and Segment Anything Model 2}

The \textit{Segment Anything Model} (SAM)~\citep{sam} and its successor, the \textit{Segment Anything Model 2} (SAM 2)~\citep{ravi2024sam2}, introduced by Meta, are significant advancements in image and video segmentation. These models aim to provide a unified framework for segmentation tasks, drawing inspiration from foundational models in NLP and CV. 
SAM focuses on image segmentation using promptable tasks to generate valid masks based on user-defined prompts. SAM 2 extends these capabilities to video segmentation, addressing challenges such as object motion and deformation.

\smallskip
\noindent
\textbf{Model.} SAM's architecture includes an image encoder for embeddings, a prompt encoder, and a mask decoder to integrate inputs and predict masks.
SAM 2 enhances SAM’s architecture with video processing capabilities. It introduces a temporal component for handling video frames, generating spatio-temporal masks (masklets) to track objects across frames.

\smallskip
\noindent
\textbf{Data.} SAM is trained on the SA-1B dataset, containing over 1 billion masks from 11 million images, ensuring robust generalization.
SAM 2 extends the dataset to include annotated video sequences, allowing it to learn from dynamic scenes and temporal changes.

\smallskip
\noindent
\textbf{Task.} SAM's promptable segmentation task generates masks based on prompts that define target objects within an image, producing plausible masks even for ambiguous prompts.
SAM 2 expands this task to video data, generating masklets that track objects across frames, maintaining accuracy despite object motion and varying conditions.

In summary, SAM addresses image segmentation, while SAM 2 extends capabilities to video segmentation. 
For comprehensive details, refer to the primary publications~\citep{sam,ravi2024sam2} and relevant surveys~\citep{zhang2023comprehensive}.

\subsection{Segment Anything in Medical Images}

Leveraging the foundational pre-trained models of SAM, various studies have investigated its effectiveness in diverse zero-shot medical imaging segmentation (MIS) scenarios. For instance, \citet{ji2023sam} performed an extensive evaluation of SAM in the \textit{everything} mode for segmenting lesion regions in different anatomical structures (e.g., brain, lung, and liver) and imaging modalities (CT and MRI).

Further, \citet{ji2023segment} analyzed SAM's performance in specific medical fields, such as optical disc and cup, polyp, and skin lesion segmentation. They used both the automatic \textit{everything} mode and the manual \textit{prompt} mode, employing points and bounding boxes as prompts.

In the context of MRI brain extraction, \citet{mohapatra2023brain} compared SAM's performance to the well-known \textit{Brain Extraction Tool} (BET) from the \textit{FMRIB Software Library}. 
Additionally, \citet{deng2023segment} evaluated SAM's capabilities in digital pathology segmentation tasks, including the segmentation of tumor, non-tumor tissue, and cell nuclei in high-resolution whole-slide images. \citet{zhou2023can} applied SAM to polyp segmentation tasks using five benchmark datasets under the \textit{everything} setting.

Recently, multiple studies have rigorously assessed SAM on over ten publicly available MIS datasets or tasks~\citep{he2023accuracy, mazurowski2023segment, ma2023segment, wu2023medical, huang2023segment, zhang2023customized}. Moreover, \citet{liu2023samm} integrated SAM with the \textit{3D Slicer} software to facilitate the design, evaluation, and application of SAM in medical imaging segmentation.

Quantitative experimental results from these studies suggest that SAM's zero-shot performance is generally moderate and varies across different datasets and tasks. Specifically:
1. Using the \textit{prompt} mode instead of the \textit{everything} mode, SAM can exceed state-of-the-art (SOTA) performance in tasks involving large objects, smaller quantities, and well-defined boundaries, especially with dense human feedback.
2. However, a significant performance gap exists between SAM and SOTA methods in tasks involving dense and amorphous object segmentation.
3. It is also important to note that most deep learning-based MIS methods require retraining from scratch for specific subtasks, and SAM-based methods are primarily limited to 2D images.

\section{Experiments and Results}

In this study, we primarily aim to explore whether annotations made on 2D medical slices using SAM 2 can be extended to entire 3D slices. 
If feasible, this could significantly reduce the annotation cost for physicians. 
To ensure the generalizability of our experimental results, we have selected two datasets: Brats2020~\citep{menze2014multimodal}  and the medical segmentation decathlon (MSD)~\citep{antonelli2022medical}. 
These datasets include MRI and CT images and encompass various commonly used medical organs and lesions.

\subsection{Datasets}

In this work, we primarily experiment with SAM 2 on two datasets: BraTS2020 and MSD.

The BraTS2020 dataset is part of the Brain Tumor Segmentation Challenge, focusing on the segmentation of gliomas in pre-operative MRI scans. It includes multimodal scans available as NIfTI files, covering native (T1), post-contrast T1-weighted (T1Gd), T2-weighted (T2), and T2 Fluid Attenuated Inversion Recovery (T2-FLAIR) volumes. We chose T2-FLAIR as our input 3D image modality because it is particularly effective in highlighting differences between normal and abnormal brain tissue, making it ideal for identifying and segmenting brain tumors. Our target is to segment the entire tumor area, including the enhancing tumor, the peritumoral edema, and the necrotic core.

The Medical Segmentation Decathlon (MSD) dataset is another significant resource designed to evaluate generalizable algorithms across various medical image segmentation tasks. It includes diverse imaging modalities and anatomical structures, such as MRI and CT scans of different organs. We utilized several tasks from MSD to segment specific organs:
Task03\_Liver: Focuses on segmenting liver structures in CT images, identifying the liver.
Task06\_Lung: Aims to segment lung regions in CT scans.
Task07\_Pancreas: Involves segmenting the pancreas in CT images.
Task09\_Spleen: Targets the segmentation of the spleen in CT scans.

These tasks help develop and benchmark robust segmentation algorithms across different medical imaging modalities.

\subsection{Evaluation Metrics}
In our experiments, we utilize the Dice coefficient and the 95\% Hausdorff distance (HD) as evaluation metrics:

\begin{itemize}
\item \textbf{Dice Coefficient}~\citep{dice1945measures}:
The Dice coefficient is a measure of similarity between two sets, often used to gauge the accuracy of segmentation. It is calculated as follows:

\begin{equation}
\rm{Dice}(X,Y)=\frac{
2 \cdot \Vert X \cap Y \Vert_1}
{\Vert X \Vert_1 + \Vert Y \Vert_1}.
\end{equation}

A higher Dice coefficient indicates a greater overlap between the predicted segmentation and the ground truth, reflecting a more accurate segmentation result.

\item \textbf{Normalized Surface Dice (NSD)}~\citep{deepmind2018surface}:
The Normalized Surface Dice (NSD) is a metric that quantifies the similarity between two sets of points, typically surfaces in a three-dimensional space. The NSD is defined as:

\begin{equation}
\text{NSD}(X, Y) = \frac{|\{x \in X \mid d(x, Y) \leq \delta\} \cap \{y \in Y \mid d(y, X) \leq \delta\}|}{|X| + |Y|},
\end{equation}

where $d(a, B)$ represents the minimum Euclidean distance from point $a$ to set $B$, and $\delta$ is a predefined distance threshold. This metric effectively measures the proportion of surface points from one set within a specified distance $\delta$ of the other set's surface points, normalized by the total number of surface points in both sets. The NSD score ranges from 0 to 1, where a score closer to 1 indicates higher similarity between the two surfaces.
\end{itemize}

\subsection{Main Results}
\begin{figure}[htb!]
    \centering
    \includegraphics[width=0.7\linewidth]{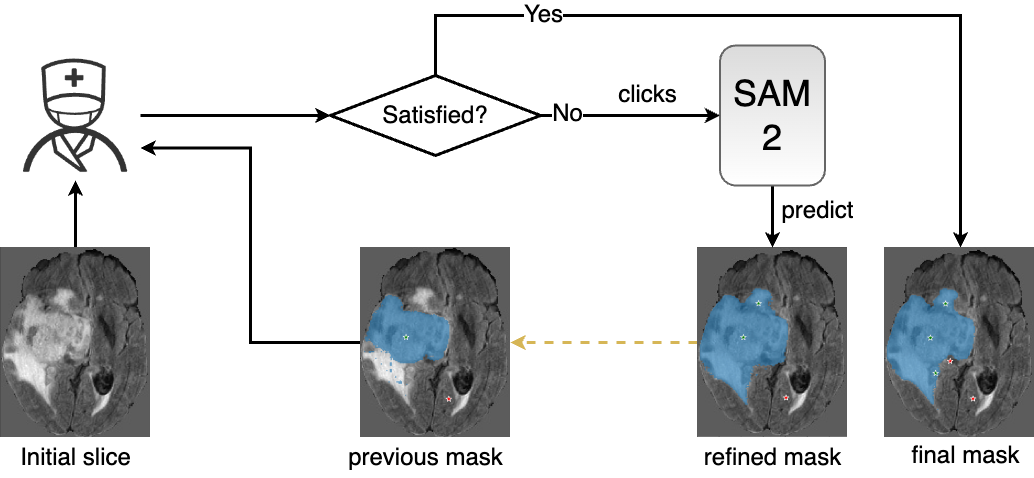}
    \caption{Interactive segmentation on a slice with SAM 2.}
    \label{fig:interactive}
\end{figure}

In this section, we present the performance of SAM 2 under different datasets and different settings.
Our experiment loads the `sam2\_hiera\_large' checkpoint and mainly focuses on two settings. 
The first involves multiple rounds of interaction on a single slice before propagating to the entire 3D image (shown in Fig.\ref{fig:interactive}). 
The second setting involves annotating a single slice and then propagating it to the entire 3D image.

\subsubsection{Compared with state-of-the-art methods}

We compare the performance of SAM 2 with several state-of-the-art 3D interactive segmentation methods, including DeepIGeoS, InterCNN, IteR-MRL, and MECCA, on the BraTS2020, Spleen, and Liver datasets. As shown in Table \ref{tab:inter_method_comparison} and \ref{tab:msd_comparison}, SAM 2 was tested under different configurations: with five interactive clicks (5 clicks) and a single ground truth mask (1 mask), both with and without focusing on the salient area.

To be noticed, these state-of-the-art 3D interactive segmentation methods are trained in resized image schema as 3D images in their original size are always too large to be loaded for training. 
Also, the resized schema needs no extra process for human feedback, as the whole image can be input into the networks.
The results indicate that while SAM 2 generally lags behind the best-performing methods for BraTS and Spleen, it shows a significant improvement in the Liver dataset. Notably, using the "1 mask" setting in the salient area for the Liver dataset, SAM 2 surpasses the best results by a considerable margin.

SAM 2 is also evaluated against several supervised methods, including nnUNet, DiNTS, Swin UNETR, and Universal Model, across different organ segmentation tasks: Spleen, Liver, Lung, and Pancreas, as presented in Table \ref{tab:msd_comparison}. 
Different from the resize schema, which is commonly used in 3D interactive medical image segmentation, these methods are trained with patches.
Patch-based training can keep the origin resolution without losing any details, which ensures high segmentation accuracy.

To be noticed the results of the supervised method are obtained from the MSD public leaderboard.
The zero-shot method based on SAM 2 was tested on the training set. Since it was trained on a natural image dataset, there is no risk of data leakage. We must acknowledge that there may be slight differences in the distribution of the test dataset. However, for the SAM 2-based algorithm, all datasets used are unseen, and we believe this difference is negligible.

The results demonstrate that SAM 2, in its various configurations, achieves competitive performance. 
we can see that there is a difference of 10.5\% to 64.81\% between SAM 2 (5 clicks) and SOTA. The difference for SAM 2 (1 mask) is relatively smaller, ranging from $3.29\%$ to $57.84\%$.
Particularly, when tested on salient areas, SAM 2 performs comparably to the best results for Spleen and Liver segmentation. 
However, its performance varies more significantly for Lung and Pancreas segmentation tasks.

Overall, as shown in Fig.\ref{fig:compare_mean_dice}, we have averaged these results for a clear comparison.
the experimental results suggest that SAM 2 still has a gap with supervised methods and can narrow the gap in specific settings and organ types. 
Further optimization and refinement of the medical 3D images is necessary.

\begin{table}[]
\caption{Comparison with 3D interactive medical image segmentation methods on BraTS2020, Spleen, and Liver segmentation tasks. ``SAM 2 (5 clicks)'' refers to interactively clicking on five points on the central 2D image using SAM, one point
per round, to generate 2D slice annotations, which are then propagated to the 3D image. ``SAM 2 (1 mask)'' indicates providing SAM 2 with the ground truth mask of the central 2D image, which is then propagated to the 3D image. ``Salient area'' refers to results tested only on slices with more than 256 foreground points. The symbols in the following table represent the same meaning.
We use bold to indicate the best result.}
\centering
\resizebox{0.6\textwidth}{!}{%
\begin{tabular}{@{}l|c|c|c@{}}
\toprule
\textbf{Method}                 & \textbf{BraTS}                & \textbf{Spleen}               & \textbf{Liver}             \\ \midrule
DeepIGeoS~\citep{deepigeos}                      & 88.54                         & 91.97                         & 48.57                      \\
InterCNN~\citep{bredell2018iterative}                        & 88.39                         & 93.52                         & 59.92                      \\
IteR-MRL~\citep{liao2020iteratively}                      & 89.22                         & 91.50                         & 62.29                      \\
MECCA~\citep{shen2023interactive}                           & \textbf{91.02}                         & \textbf{94.96}                         & 71.46                    \\ \midrule
SAM 2 (5 clicks)                & 75.52                         & 79.59                         & 81.32                      \\
Compared with the best results  & \cellcolor{orange!30}-17.03\% & \cellcolor{orange!30}-16.19\% & \cellcolor{blue!30}13.80\% \\
SAM 2 (1 mask)                  & 81.29                         & 82.77                         & 90.18                      \\
Compared with the best results  & \cellcolor{orange!30}-10.69\% & \cellcolor{orange!30}-12.84\% & \cellcolor{blue!30}26.20\% \\
SAM 2 (5 clicks) (salient area) & 81.12                         & 92.98                         & 84.85                      \\
Compared with the best results  & \cellcolor{orange!30}-10.88\% & \cellcolor{orange!30}-2.09\%  & \cellcolor{blue!30}18.74\% \\
SAM 2 (1 mask) (salient area)   & 87.17                         & 94.41                         & \textbf{92.33}                      \\
Compared with the best results  & \cellcolor{orange!30}-4.23\%  & \cellcolor{orange!30}-0.58\%  & \cellcolor{blue!30}29.21\% \\ \bottomrule
\end{tabular}}
\label{tab:inter_method_comparison}
\end{table}

\begin{table}[]
\caption{Comparison with supervised methods for various organs. }
\resizebox{\textwidth}{!}{%
\begin{tabular}{@{}lcccccccc@{}}
\toprule
\multirow{2}{*}{\textbf{Method}}                                & \multicolumn{2}{|c|}{\textbf{Spleen}}                                    & \multicolumn{2}{c|}{\textbf{Liver}}                                     & \multicolumn{2}{c|}{\textbf{Lung}}                                      & \multicolumn{2}{c}{\textbf{Pancreas}}                           \\ \cmidrule(l){2-9} 
                                                                & \multicolumn{1}{|c|}{\textbf{Dice}} & \multicolumn{1}{c|}{\textbf{NSD}} & \multicolumn{1}{c|}{\textbf{Dice}} & \multicolumn{1}{c|}{\textbf{NSD}} & \multicolumn{1}{c|}{\textbf{Dice}} & \multicolumn{1}{c|}{\textbf{NSD}} & \multicolumn{1}{c|}{\textbf{Dice}} & \textbf{NSD}               \\ \midrule
\multicolumn{1}{l|}{nnUNet ~\citep{nnunet}}                     & \multicolumn{1}{c|}{\textbf{97.43}}         & \multicolumn{1}{c|}{\textbf{99.89}}        & \multicolumn{1}{c|}{\textbf{95.75}}         & \multicolumn{1}{c|}{98.55}        & \multicolumn{1}{c|}{73.97}         & \multicolumn{1}{c|}{76.02}        & \multicolumn{1}{c|}{81.64}         & 96.14                      \\
\multicolumn{1}{l|}{DiNTS \citep{he2023swinunetr}}              & \multicolumn{1}{c|}{96.98}         & \multicolumn{1}{c|}{99.83}        & \multicolumn{1}{c|}{95.35}         & \multicolumn{1}{c|}{\textbf{98.69}}        & \multicolumn{1}{c|}{74.75}         & \multicolumn{1}{c|}{77.53}        & \multicolumn{1}{c|}{81.02}         & 96.26                      \\
\multicolumn{1}{l|}{Swin UNETR ~\citep{tang2024efficient}}      & \multicolumn{1}{c|}{96.99}         & \multicolumn{1}{c|}{99.84}        & \multicolumn{1}{c|}{95.35}         & \multicolumn{1}{c|}{98.34}        & \multicolumn{1}{c|}{\textbf{76.60}}          & \multicolumn{1}{c|}{77.40}         & \multicolumn{1}{c|}{81.85}         & 96.57                      \\
\multicolumn{1}{l|}{Universal Model \citep{liu2024universal}}   & \multicolumn{1}{c|}{97.27}         & \multicolumn{1}{c|}{99.87}        & \multicolumn{1}{c|}{95.42}         & \multicolumn{1}{c|}{98.18}        & \multicolumn{1}{c|}{80.01}         & \multicolumn{1}{c|}{\textbf{81.25}}        & \multicolumn{1}{c|}{\textbf{82.84}}         & \textbf{96.65}                      \\ \midrule
\multicolumn{1}{l|}{SAM 2 (5 clicks)}                           & \multicolumn{1}{c|}{79.59}         & \multicolumn{1}{c|}{75.63}        & \multicolumn{1}{c|}{81.32}         & \multicolumn{1}{c|}{50.47}        & \multicolumn{1}{c|}{71.61}         & \multicolumn{1}{c|}{68.99}        & \multicolumn{1}{c|}{44.73}         & 34.01                      \\
\multicolumn{1}{l|}{Compared with the best results}             & \multicolumn{1}{c|}{\cellcolor{orange!30}-18.31\%}      & \multicolumn{1}{c|}{\cellcolor{orange!30}-24.29\%}     & \multicolumn{1}{c|}{\cellcolor{orange!30}-15.07\%}      & \multicolumn{1}{c|}{\cellcolor{orange!30}-48.86\%}     & \multicolumn{1}{c|}{\cellcolor{orange!30}-10.50\%}      & \multicolumn{1}{c|}{\cellcolor{orange!30}-15.09\%}     & \multicolumn{1}{c|}{\cellcolor{orange!30}-46.00\%}      & \cellcolor{orange!30}-64.81\%                   \\ \midrule
\multicolumn{1}{l|}{SAM 2 (1 mask)}                             & \multicolumn{1}{c|}{82.77}         & \multicolumn{1}{c|}{79.35}        & \multicolumn{1}{c|}{90.18}         & \multicolumn{1}{c|}{61.29}        & \multicolumn{1}{c|}{77.38}         & \multicolumn{1}{c|}{74.97}        & \multicolumn{1}{c|}{51.48}         & 40.75                      \\
\multicolumn{1}{l|}{Compared with the best results}             & \multicolumn{1}{c|}{\cellcolor{orange!30}-15.05\%}      & \multicolumn{1}{c|}{\cellcolor{orange!30}-20.56\%}     & \multicolumn{1}{c|}{\cellcolor{orange!30}-5.82\%}       & \multicolumn{1}{c|}{\cellcolor{orange!30}-37.90\%}     & \multicolumn{1}{c|}{\cellcolor{orange!30}-3.29\%}       & \multicolumn{1}{c|}{\cellcolor{orange!30}-7.73\%}      & \multicolumn{1}{c|}{\cellcolor{orange!30}-37.86\%}      & \cellcolor{orange!30}-57.84\%                   \\ \midrule
\multicolumn{1}{l|}{SAM 2 (5 clicks)(salient area)}             & \multicolumn{1}{c|}{92.98}         & \multicolumn{1}{c|}{89.71}        & \multicolumn{1}{c|}{84.85}         & \multicolumn{1}{c|}{52.69}        & \multicolumn{1}{c|}{83.93}         & \multicolumn{1}{c|}{78.68}        & \multicolumn{1}{c|}{51.45}         & 40.82                      \\
\multicolumn{1}{l|}{Compared with the best results}             & \multicolumn{1}{c|}{\cellcolor{orange!30}-4.57\%}       & \multicolumn{1}{c|}{\cellcolor{orange!30}-10.19\%}     & \multicolumn{1}{c|}{\cellcolor{orange!30}-11.38\%}      & \multicolumn{1}{c|}{\cellcolor{orange!30}-46.61\%}     & \multicolumn{1}{c|}{\cellcolor{blue!30}4.90\%}        & \multicolumn{1}{c|}{\cellcolor{orange!30}-3.16\%}      & \multicolumn{1}{c|}{\cellcolor{orange!30}-37.89\%}      & \cellcolor{orange!30}-57.77\%                   \\ \midrule
\multicolumn{1}{l|}{SAM 2 (1 mask)(salient area)}               & \multicolumn{1}{c|}{94.41}         & \multicolumn{1}{c|}{92.46}        & \multicolumn{1}{c|}{92.33}         & \multicolumn{1}{c|}{63.13}        & \multicolumn{1}{c|}{\textbf{87.48}}         & \multicolumn{1}{c|}{\textbf{82.39}}        & \multicolumn{1}{c|}{61.04}         & 50.65                      \\
\multicolumn{1}{l|}{Compared with the best results}             & \multicolumn{1}{c|}{\cellcolor{orange!30}-3.10\%}       & \multicolumn{1}{c|}{\cellcolor{orange!30}-7.44\%}      & \multicolumn{1}{c|}{\cellcolor{orange!30}-3.57\%}       & \multicolumn{1}{c|}{\cellcolor{orange!30}-36.03\%}     & \multicolumn{1}{c|}{\cellcolor{blue!30}9.34\%}        & \multicolumn{1}{c|}{\cellcolor{blue!30}1.40\%}       & \multicolumn{1}{c|}{\cellcolor{orange!30}-26.32\%}      & \cellcolor{orange!30}-47.59\%                   \\ \bottomrule
\end{tabular}%
}

\label{tab:msd_comparison}
\end{table}

\subsubsection{Statistics of improvement brought about by interaction}
SAM 2, benefiting from the SA-V dataset, which comprises $50.9$K videos and $642.6$K masklets, and its carefully designed architecture, demonstrates robust zero-shot inference capabilities on natural images. 
We evaluated the performance of SAM 2 on the BraTS2020 benchmark and compared it with 3D interactive medical segmentation algorithms.

The algorithms DeepIGeoS, InterCNN, IteR-MRL, and MECCA adopt direct clicks on 3D medical images over five rounds, with 25 interaction points provided in the first round and 5 additional points in each subsequent round. 
For SAM 2, we employed a similar setup; however, interactions were conducted on 2D slices with only one point per round, which SAM 2 then propagates to the entire 3D image.

To fairly compare the algorithms' utilization of interactive feedback, we selected the Dice Score Growth per Added Point, which is the increase in dice score during a round divided by the number of new points added. 
As shown in Figure \ref{fig:growth_per_added_point}, SAM 2's efficiency in utilizing interactive feedback significantly surpasses that of other algorithms. 
This demonstrates that SAM 2 possesses strong refinement capabilities based on interactions in the medical imaging domain.

\begin{figure}[htb!]
    \centering
    \includegraphics[width=0.8\linewidth]{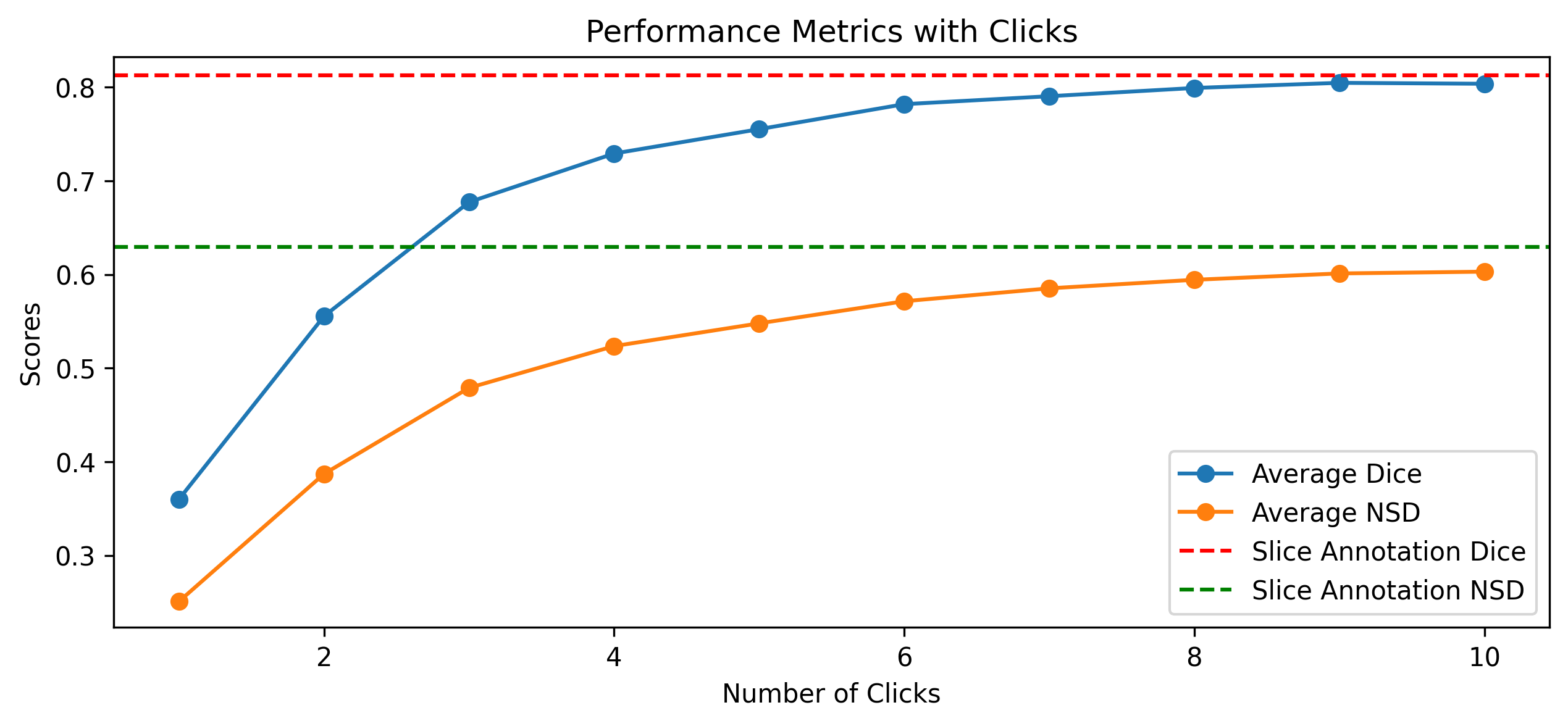}
    \caption{SAM 2 with different iterative steps on Brats2020 benchmark.}
    \label{fig:clicks_num}
\end{figure}

Additionally, we assessed the discrepancy between slice annotations obtained through multiple rounds of interactive clicks and the slice ground truth. 
As shown in Figure \ref{fig:clicks_num}, performance gradually improves and approaches the ground truth with an increasing number of clicks. 
This further validates the feasibility of interactive 2D slice segmentation followed by propagation using SAM 2.

\section{Conclusion}
In this paper, we investigated the application of the Segment Anything Model 2 (SAM 2) for zero-shot 3D medical image segmentation. 
By leveraging its ability to propagate annotations from a single 2D slice to an entire 3D volume, SAM 2 addresses the limitation of traditional 2D trained models that they struggle with 3D medical images because they can't efficiently use the features and annotations from one slice across other slices.
Our experiments on the BraTS2020 and MSD datasets reveal that SAM 2, while not yet matching the performance of specialized supervised methods, shows promising results in specific settings and organ types. 
The efficiency of SAM 2 in utilizing interactive feedback surpasses that of other 3D interactive segmentation algorithms, demonstrating its potential to significantly reduce the annotation workload for medical professionals. 
However, further optimization and refinement are necessary to enhance its performance and generalizability. 
This empirical study lays the groundwork for future research into leveraging advanced models like SAM 2 to revolutionize 3D medical image segmentation, ultimately improving clinical workflows and patient outcomes.

\clearpage
\newpage

\bibliography{main}

\clearpage
\newpage

\begin{appendices}
\counterwithin{figure}{section}
\section{Visualization Results}

\begin{figure}[htb!]
    \centering
    \includegraphics[width=\linewidth]{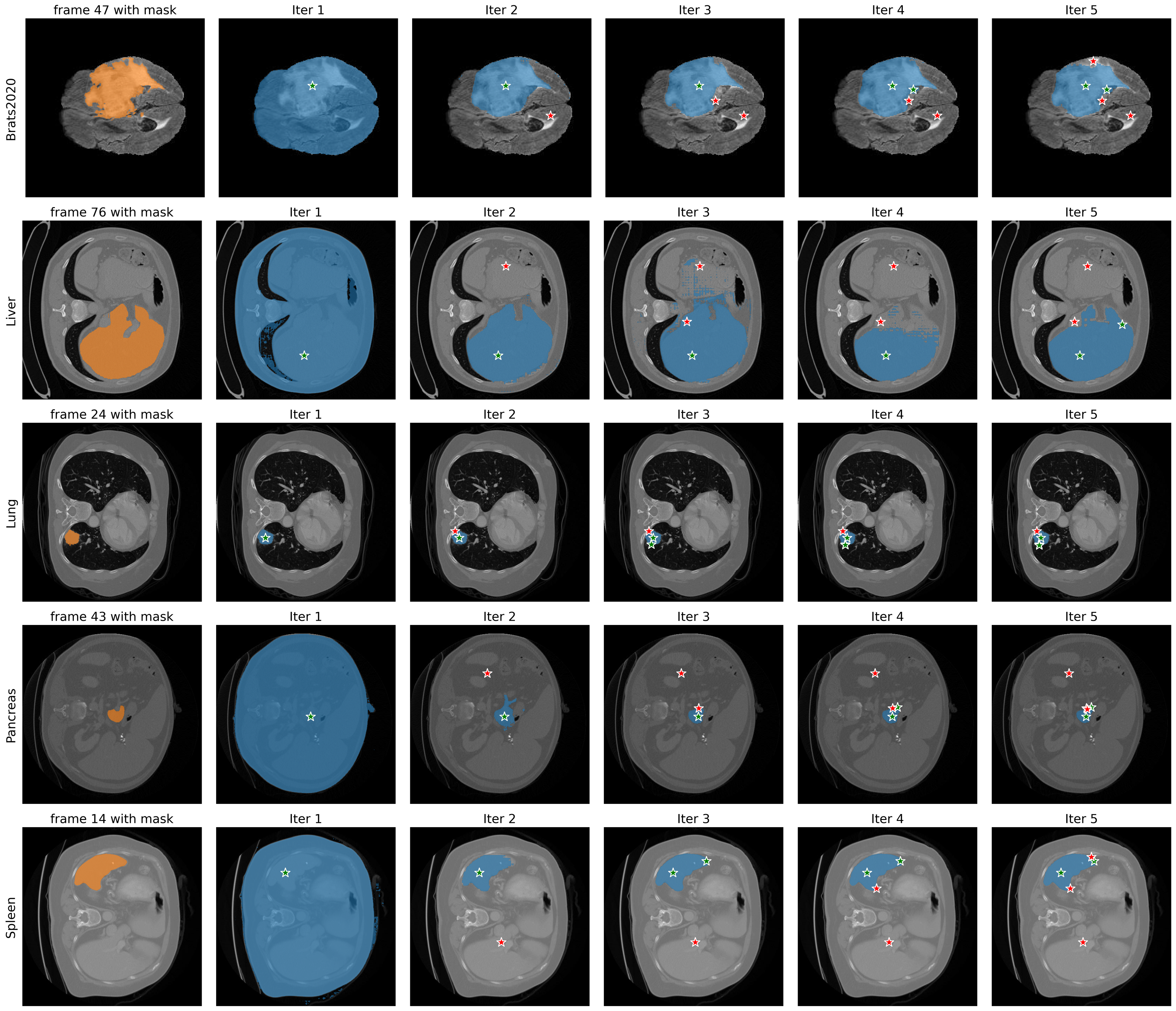}
    \caption{SAM 2 segmentation 2D slices with 5 interactive clicks feedback. We show the ground truth mask as orange and the predicted mask as blue. We show foreground clicks in blue and background clicks in orange.}
    \label{fig:clicks}
\end{figure}

\begin{figure}[htb!]
    \centering
    \includegraphics[width=\linewidth]{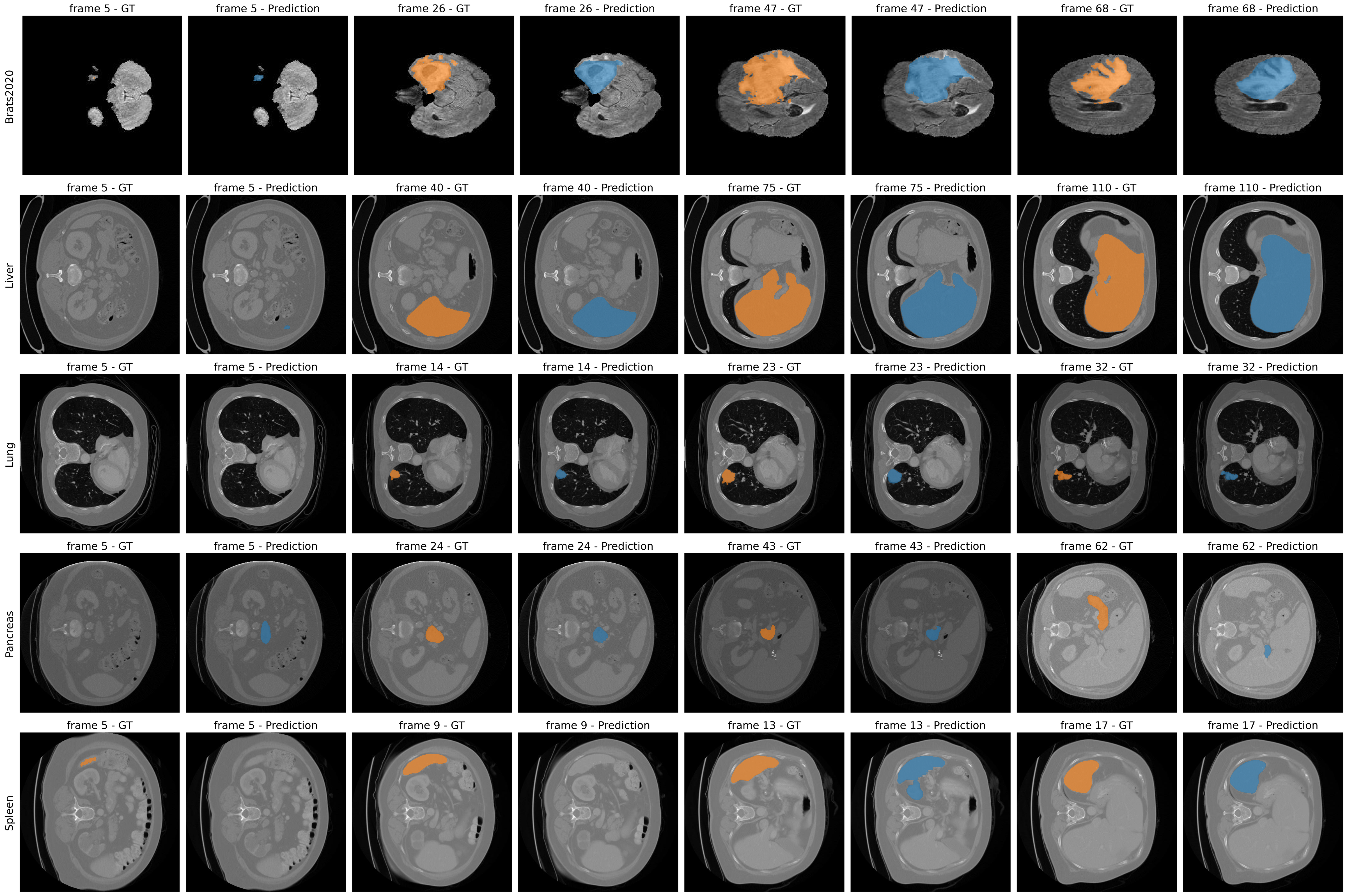}
    \caption{Propagation: We show the ground truth mask as orange and the predicted mask as blue.}
    \label{fig:propa}
\end{figure}

To further qualitatively study the accuracy of SAM 2 on medical images, we visualized the multi-round 2D slice interactive segmentation for brain tumors and different organs in Fig.\ref{fig:clicks}. Each row represents a different dataset, with the first column showing the Ground Truth, followed by the results of each subsequent round. It can be observed that SAM 2 effectively refines the results gradually, producing masks that closely resemble the ground truth.

Furthermore, using the interactive masks obtained from the fifth round, we applied SAM 2 to propagate the segmentation across the entire 3D image. The results on different datasets are shown in Fig.\ref{fig:propa}. Each row represents a different dataset, with odd-numbered columns showing the ground truth of corresponding slices, followed by the predicted masks. Significant differences can be seen between slices of the 3D image, indicating substantial morphological variations. In most cases, SAM 2 demonstrates good performance, although there are some failures, such as the last column in the spleen segmentation task, where the target regions in earlier slices were not identified by SAM 2.

These visual results demonstrate the zero-shot capability of SAM 2, which can achieve relatively accurate segmentation on medical images despite the significant differences from natural images. However, the precision of the segmentation still requires further improvement.

\end{appendices}

\end{document}